\documentclass[letterpaper, 10 pt, conference]{ieeeconf}  

\IEEEoverridecommandlockouts                              
\usepackage[utf8]{inputenc}
\usepackage[T1]{fontenc}
\usepackage{subcaption}
\usepackage{tikz}
\usepackage{svg}
\usepackage{comment}
\usepackage{pifont}

\usepackage{multirow} 
\usepackage{cite}
\usepackage{graphicx}

\usepackage{soul}

\usepackage{amsmath}

\usepackage{xcolor,colortbl}

\definecolor{rblue}{rgb}{0,0.5,1}
\definecolor{awesome}{rgb}{1.0, 0.13, 0.32}
\definecolor{hollywoodcerise}{rgb}{0.96, 0.0, 0.63}
\definecolor{lasallegreen}{rgb}{0.03, 0.47, 0.19}
\definecolor{hanpurple}{rgb}{0.32, 0.09, 0.98}
\definecolor{green(pigment)}{rgb}{0.0, 0.65, 0.31}
\definecolor{ourgreen}{RGB}{235,250,235}
\definecolor{fisheye_orange}{RGB}{197,153,13}
\definecolor{go_c}{RGB}{50,255,255}
\definecolor{vehicle_c}{RGB}{100,150,245}
\definecolor{parking_c}{RGB}{255,150,255}
\definecolor{fence_c}{RGB}{255,120,5}
\definecolor{pedestrian_c}{RGB}{255,30,30}
\definecolor{sign_c}{RGB}{255,30,30}
\definecolor{cyclist_c}{RGB}{255,40,200}
\definecolor{tl_c}{RGB}{255,150,0}
\definecolor{pole_c}{RGB}{255,240,150}
\definecolor{cc_c}{RGB}{255,120,50}
\definecolor{bicycle_c}{RGB}{100,230,240}
\definecolor{building_c}{RGB}{255,200,0}
\definecolor{vegetation_c}{RGB}{0,175,0}
\definecolor{tt_c}{RGB}{135,60,0}
\definecolor{road_c}{RGB}{255,0,255}
\definecolor{walkable_c}{RGB}{75,0,75}

\definecolor{truck_c}{RGB}{80, 30, 180}
\definecolor{ov_c}{RGB}{0, 0, 255}
\definecolor{og_c}{RGB}{175, 0, 75}
\definecolor{os_c}{RGB}{255, 150, 0}
\definecolor{mygray}{gray}{.9}
\usepackage{booktabs}

\makeatletter
\let\NAT@parse\undefined
\makeatother
\usepackage[pagebackref=false,breaklinks=true,colorlinks,bookmarks=false]{hyperref}
\hypersetup{colorlinks=true,linkcolor={red},citecolor={hanpurple},urlcolor={magenta}}

\newcommand{\pinhole}{\includegraphics[height=1em]{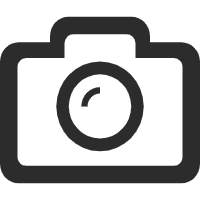}}
\newcommand{\lidar}{\includegraphics[height=1em]{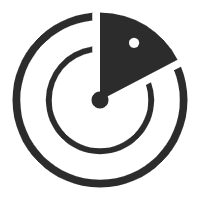}}

\newcommand{\real}{\includegraphics[height=1em]{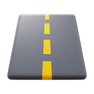}}
\newcommand{\syn}{\includegraphics[height=1em]{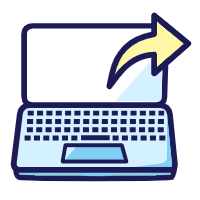}}
\newcommand{\fisheye}{\includegraphics[height=1em]{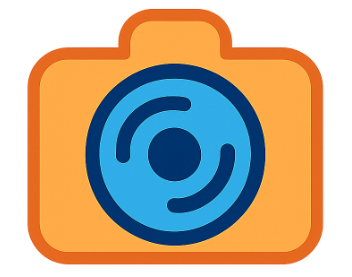}}

\title{\LARGE \bf
OccTrack360: 4D Panoptic Occupancy Tracking from Surround-View Fisheye Cameras}

\author{Yongzhi Lin$^{1,*}$, Kai Luo$^{1,*}$, Yuanfan Zheng$^{1}$, Hao Shi$^{2,3}$, Mengfei Duan$^{1}$, Yang Liu$^{1}$, and Kailun Yang$^{1,\dag}$
\thanks{This work was supported in part by the National Natural Science Foundation of China (Grant No. 62473139), in part by the Hunan Provincial Research and Development Project (Grant No. 2025QK3019), and in part by the State Key Laboratory of Autonomous Intelligent Unmanned Systems (the opening project number ZZKF2025-2-10).}%
\thanks{$^{1}$The authors are with the School of Artificial Intelligence and Robotics and the National Engineering Research Center of Robot Visual Perception and Control Technology, Hunan University, Changsha, China (email: kailun.yang@hnu.edu.cn).}%
\thanks{$^{2}$The author is with Ant Group, Hangzhou, China.}%
\thanks{$^{3}$The author is also with the State Key Laboratory of Extreme Photonics and Instrumentation, Zhejiang University, Hangzhou, China.}%
\thanks{$^{*}$Equal contribution.}
\thanks{$^{\dag}$Corresponding author: Kailun Yang.}
}

\begin{document}

\maketitle
\thispagestyle{empty}
\pagestyle{empty}

\begin{abstract}
Understanding dynamic 3D environments in a spatially continuous and temporally consistent manner is fundamental for robotics and autonomous driving. While recent advances in occupancy prediction provide a unified representation of scene geometry and semantics, progress in 4D panoptic occupancy tracking remains limited by the lack of benchmarks that support surround-view fisheye sensing, long temporal sequences, and instance-level voxel tracking. To address this gap, we present \textbf{OccTrack360}, a new benchmark for 4D panoptic occupancy tracking from surround-view fisheye cameras. OccTrack360 provides substantially longer and more diverse sequences ($174{\sim}2234$ frames) than prior benchmarks, together with principled voxel visibility annotations, including an all-direction occlusion mask and an MEI-based fisheye field-of-view mask. To establish a strong fisheye-oriented baseline, we further propose \textbf{Focus on Sphere Occ (FoSOcc)}, a framework that addresses two core challenges in fisheye occupancy tracking: distorted spherical projection and inaccurate voxel-space localization. FoSOcc includes a \textit{Center Focusing Module (CFM)} to enhance instance-aware spatial localization through supervised focus guidance, and a \textit{Fisheye-based Enhanced Lifting (FEL)} that extends perspective lifting to fisheye imaging under the Unified Projection Model. Extensive experiments on Occ3D-Waymo and OccTrack360 show that our method improves occupancy tracking quality with notable gains on geometrically regular categories, and establishes a strong baseline for future research on surround-view fisheye 4D occupancy tracking. The benchmark and source code will be made publicly available at \url{https://github.com/YouthZest-Lin/OccTrack360}.
\end{abstract}

\section{Introduction}

A comprehensive and reliable understanding of dynamic 3D environments is a prerequisite for perception systems in robotics and autonomous driving~\cite{xu2025survey,hu2025survey}. 
Beyond recognizing objects in isolated frames, a deployable system must infer scene geometry, semantics, and object identities in a spatially continuous and temporally consistent manner, so that it can reason about both static structures and moving agents over time. 
Recent progress in camera-based occupancy prediction has made this direction increasingly practical by representing the world as dense voxels with semantic labels, offering a unified interface for downstream tasks such as planning and interaction~\cite{chen2025alocc,li2025voxdet}.

Despite these advances, progress in 4D panoptic occupancy tracking is still constrained by the lack of a suitable benchmark capturing the full spatiotemporal surroundings. 
Existing occupancy datasets and benchmarks~\cite{tian2023occ3d,li2024sscbench} primarily focus on pinhole-camera setups, limited Fields of View (FoV), or relatively short sequences, which makes them insufficient for evaluating long-term, surround-view dynamic understanding. 
In particular, they do not jointly provide: (1) wide-FoV fisheye observations, (2) temporally consistent instance-level voxel annotations, and (3) principled visibility constraints for occupancy tracking. 
As a result, current evaluations cannot fully reflect the challenges encountered in real robotic systems, where wide-FoV sensing and persistent object identities are both essential.

\begin{figure}[!t] %
    \centering
    \includegraphics[width=\columnwidth, height=5cm, keepaspectratio]{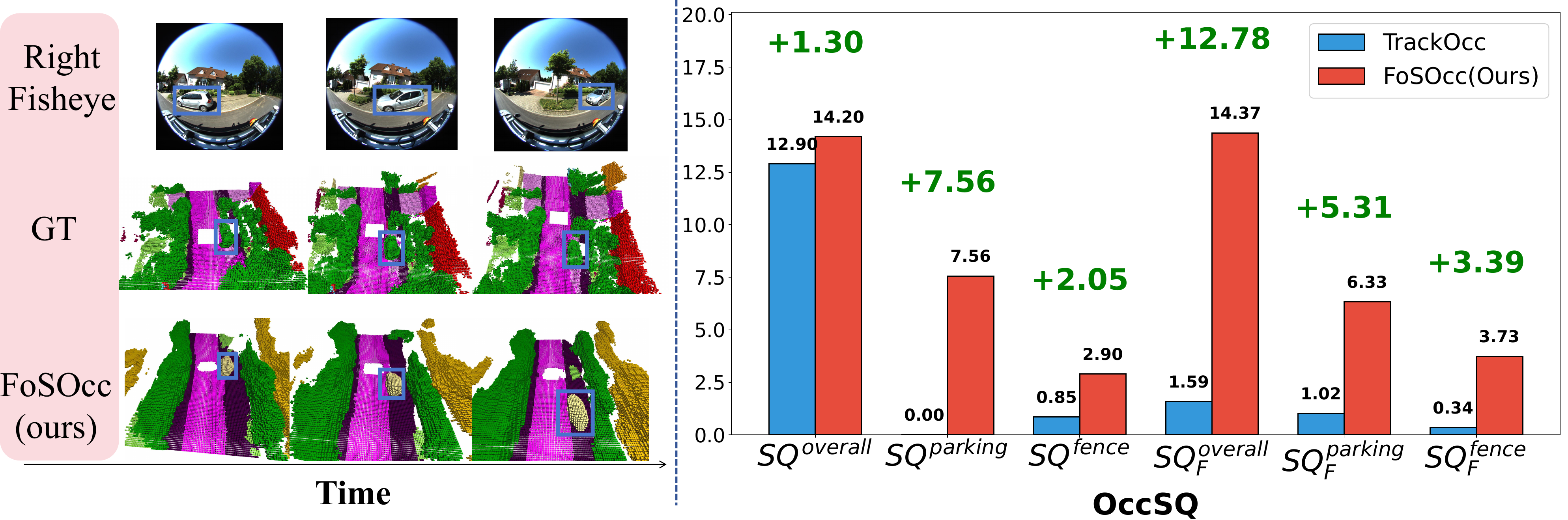}
    \vskip-2ex
    \caption{\textbf{Illustration of 4D panoptic occupancy tracking.} Objects highlighted within the blue rectangles are the same objects. Our FoSOcc leverages fisheye images as input to perform comprehensive 4D panoptic occupancy tracking.}
    \label{fig:teaser}
    \vskip-4ex
\end{figure}

To address this gap, we introduce \textbf{OccTrack360}, a new benchmark for 4D panoptic occupancy tracking from surround-view fisheye cameras. 
OccTrack360 is designed to support holistic scene understanding under realistic fisheye imaging. Compared with prior occupancy tracking benchmarks, OccTrack360 provides substantially longer and more diverse temporal sequences (ranging from $174$ to $2234$ frames), enabling meaningful evaluation of temporal consistency and long-horizon tracking behavior. 
In addition, we construct more principled supervision signals for voxel visibility: an occlusion mask that covers all directions in the voxel domain (instead of only directions intersecting occupied voxels), and a fisheye FoV mask derived under the Unified Projection Model (MEI)~\cite{mei2007single}, which explicitly indicates whether each voxel lies inside or outside the valid fisheye field of view. 
These design choices make OccTrack360 not only a larger benchmark, but also a more faithful one for evaluating camera-only 4D occupancy tracking under surround-view fisheye sensing.

\begin{figure*}[!t]
    \centering
    \begin{minipage}[c]{0.61\linewidth}
        \centering
        
        \vskip 1ex
        \resizebox{\linewidth}{!}{
            \begin{tabular}{lcccccc}
            \toprule
            \textbf{Datasets}                                 & \textbf{Year} & \textbf{Source} & \textbf{Modality} & \textbf{\#Classes} & \textbf{Voxel Size}      & \textbf{Instance Level} \\
            \midrule
            SemanticKITTI~\cite{behley2019semantickitti}& 2019          & \real& \pinhole \lidar& 28                 & 256$\times$256$\times$32 & \ding{55}\\
            Occ3D-nuScenes~\cite{tian2023occ3d}& 2023          & \real& \pinhole& 16                 & 200$\times$200$\times$16 & \ding{55}\\
            Occ3D-Waymo~\cite{tian2023occ3d}& 2023          & \syn& \pinhole& 14                 & 200$\times$200$\times$32 & \ding{55}\\
            nuScenes-Occupancy~\cite{wang2023openoccupancy}& 2023          & \real& \pinhole \lidar& 16                 & 512$\times$512$\times$40 & \ding{55}\\
            OpenOcc~\cite{sima2023_occnet}& 2023          & \real& \pinhole   \lidar& 16                 & 200$\times$200$\times$16 & \ding{55}\\
            SSCBench-KITTI360~\cite{li2024sscbench}         & 2024          & \real& \pinhole\lidar& 19                 & 256$\times$256$\times$32 & \ding{55}\\
            WoodScape~\cite{yogamani2019woodscape}                                                             & 2019          & \real& \fisheye  \lidar& 40                 & \ding{55}& \ding{51}\\
            SynWoodScape~\cite{sekkat2022synwoodscape}                                                          & 2022          & \syn& \fisheye \lidar& 22                 & \ding{55}& \ding{51}\\
            TrackOcc-Waymo~\cite{chen2025trackocc}                                                              & 2025          & \syn& \pinhole& 14                 & 200$\times$200$\times$16 & \ding{51}\\
            \rowcolor{ourgreen} OccTrack360 (Ours)& 2026          & \real& \pinhole  \fisheye& 18                 & 128$\times$128$\times$16 & \ding{51}\\
            \bottomrule
            \end{tabular}
         }
        \vspace*{3.5pt}
        \subcaption{\textbf{Dataset comparison for occupancy and fisheye perception.}}
        \label{table:dataset_comparison}

    \end{minipage}\hfill
        \begin{minipage}[c]{0.39\linewidth}
        \centering
        \vspace*{8pt} 
        \vskip-1ex
        \includegraphics[width=0.99\linewidth]{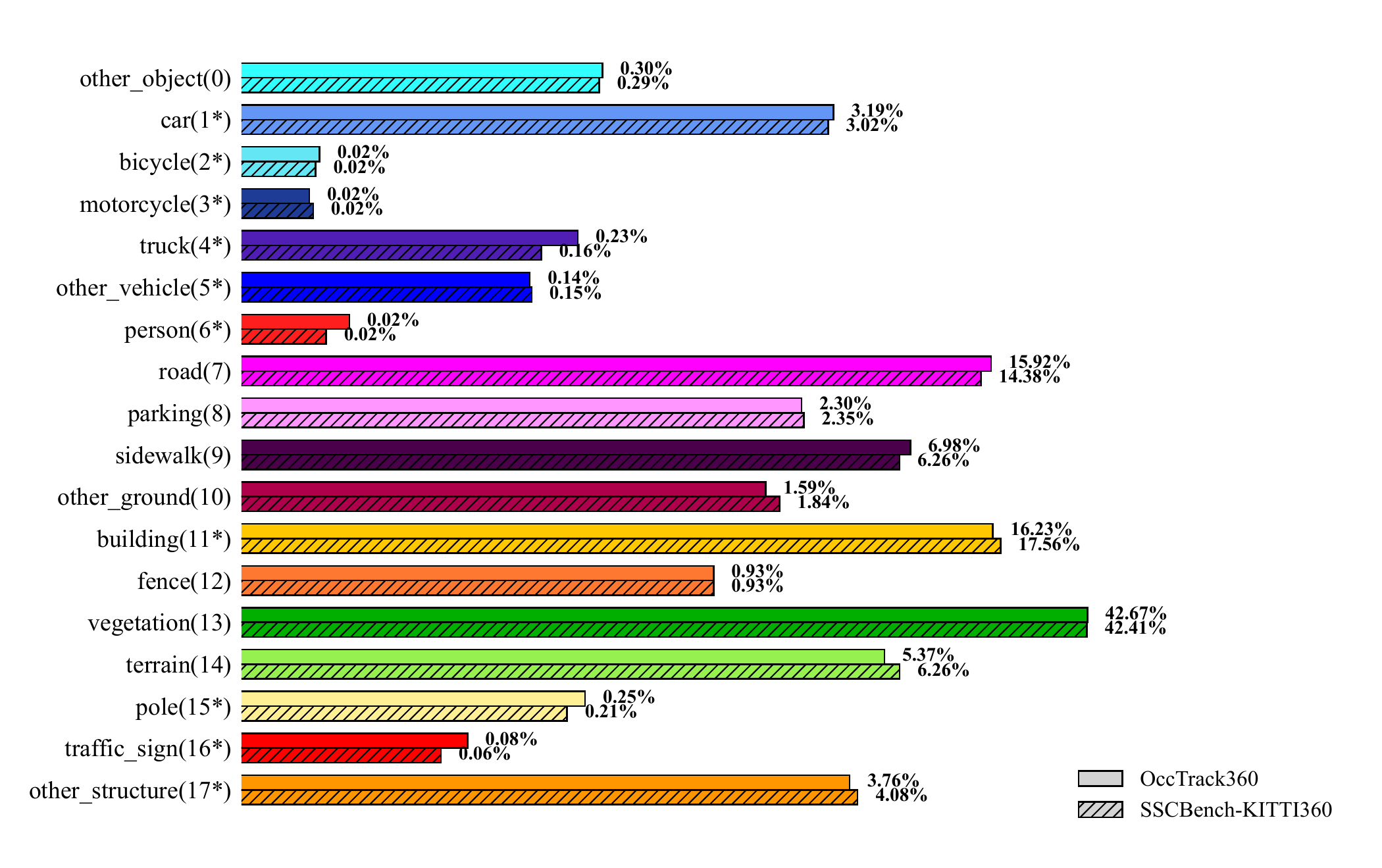}
        \vskip-2ex
        \subcaption{\textbf{Distribution of $18$ semantic classes.}}
        \label{fig:data_distribution}
    \end{minipage}
    \vskip-1ex
    \caption{OccTrack360 facilitates comprehensive 4D panoptic occupancy prediction in 18 semantic classes using both pinhole and fisheye inputs. An asterisk(*) denotes categories for which instance-level tracking labels are provided. Abbreviations: \real~(Real-World Data), \syn~(Synthetic Data), \pinhole~(Pinhole Camera), \lidar~(LiDAR), and \fisheye~(Fisheye Camera).}
    \vskip-4ex
\end{figure*}

Benchmarking on fisheye imagery also introduces unique modeling challenges. Severe radial distortion and spherical projection geometry break the assumptions behind standard perspective-camera lifting pipelines, making it difficult to accurately lift 2D image features into 3D voxel space. Meanwhile, inevitable depth estimation noise in voxel space further degrades object localization, especially for instance-level tracking. 
To tackle these issues, we propose \textbf{Focus on Sphere Occ (FoSOcc)}, a fisheye-oriented occupancy tracking framework with two key components. First, the \textit{Center Focusing Module (CFM)} introduces supervised focus cues to strengthen geometric representation and improve instance-aware spatial localization. Second, the \textit{Fisheye-based Enhanced Lifting (FEL)} extends the lift operation of LSS~\cite{philion2020lift} to fisheye/spherical imaging by incorporating the MEI camera model, enabling more geometrically consistent 2D-to-3D feature lifting under wide-FoV distortion.

We verify our approach on both Occ3D-Waymo~\cite{tian2023occ3d} and the proposed OccTrack360 benchmark. 
On Occ3D-Waymo, our CFM consistently improves occupancy segmentation quality for geometrically regular categories, yielding relative gains of $11.1\%$ on \emph{signs} and $20.7\%$ on \emph{general objects} over the baseline. 
On OccTrack360, as shown in Fig.~\ref{fig:teaser}, FoSOcc establishes strong baselines for surround-view fisheye 4D occupancy tracking and demonstrates the effectiveness of jointly modeling fisheye-aware lifting and center-focused supervision. 
These results confirm that improving benchmark design and fisheye-aware modeling must go hand in hand for advancing holistic scene understanding.

In summary, this paper makes the following contributions:
\begin{itemize}
    \item We introduce \textbf{OccTrack360}, a new benchmark for 4D panoptic occupancy tracking from surround-view fisheye cameras, with long and diverse temporal sequences, instance-level voxel annotations, and principled visibility constraints tailored to fisheye imaging.
    \item We construct an all-direction occlusion mask and an MEI-based fisheye FoV mask, providing more faithful supervision and evaluation for wide-FoV voxel reasoning in dynamic driving scenes.
    \item We propose \textbf{FoSOcc}, a fisheye-oriented 4D occupancy tracking framework with a \textit{Center Focusing Module (CFM)} for instance-aware localization and a \textit{Fisheye-based Enhanced Lifting (FEL)} for distortion-aware 2D-to-3D lifting under severe radial distortion.
    \item Extensive experiments on Occ3D-Waymo and OccTrack360 demonstrate that the proposed method is comparable to competitive alternatives and provides a strong baseline for future research on fisheye-based 4D occupancy tracking.
\end{itemize}

\section{Related Work}

\noindent\textbf{3D Occupancy Prediction and Semantic Scene Completion.}
Semantic Scene Completion (SSC) was initially popularized by datasets such as SemanticKITTI~\cite{behley2019semantickitti}, which challenge models to infer the semantic labels of both visible and occluded voxels. 
Recently, the field has transitioned toward 3D Occupancy Prediction~\cite{tian2023occ3d, li2024sscbench}, emphasizing holistic surround-view perception for autonomous driving.
A primary challenge in this domain is accurately mapping 2D image features to 3D voxel space while preserving fine-grained geometry. 
While these methodologies~\cite{gan2025gaussianocc, Chen_2025_ICCV, MinkOcc_2025} have matured, they remain predominantly optimized for pinhole camera inputs, which inherently possess a restricted FoV and create blind spots in the immediate vicinity of the ego-vehicle. 
This spatial limitation necessitates not only a more comprehensive visual field but also a sophisticated understanding of how the scene evolves over time to ensure safety in critical areas.

\begin{figure*}[!t]
    \centering
    \includegraphics[width=0.9\linewidth]{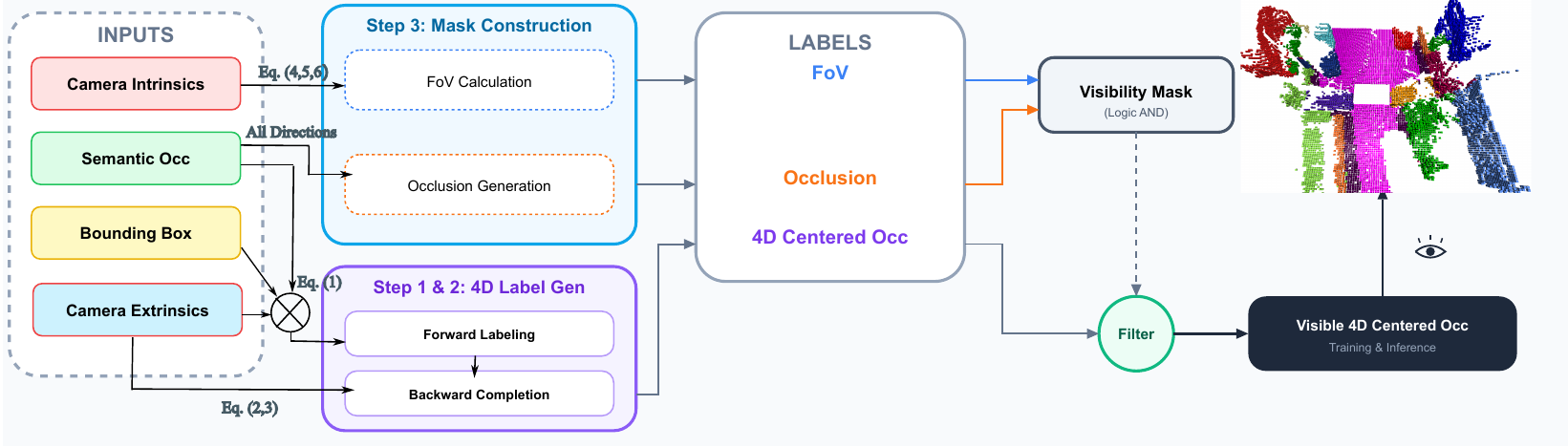}
    \caption{\textbf{Benchmark pipeline.} 
    Our benchmark integrates multiple inputs to derive the FoV mask, occlusion mask, and 4D centered occupancy labels. 
    These masks are subsequently combined to filter the visible regions, yielding the final representation used for training and inference.
    ``Gen'' denotes generation.}
    \label{fig:Fisheye_Benchmark_Pipeline}
    \vskip-4ex
\end{figure*}

\noindent\textbf{4D Instance-level Occupancy Prediction.}
Temporal integration has become a cornerstone for achieving spatial consistency in 3D occupancy, as evidenced by the shift from single-frame inference to multi-frame feature fusion~\cite{li2023fbocc, leng2025occupancy}. 
Beyond static semantic stability, recent methodologies have introduced voxel-based motion modeling and flow estimation to characterize scene dynamics~\cite{sima2023_occnet, wang2025uniocc}. 
For instance, Cam4DOcc~\cite{ma2024cam4docc} and StreamOcc~\cite{moon2025mitigating} leverage continuous temporal streams to refine geometry and predict future occupancy states, yet these approaches primarily operate at the semantic or flow level without explicit object identity association.
For object-level awareness, recent frameworks~\cite{chen2025trackocc, luz2026latent} integrate instance-aware queries to maintain ID consistency for foreground categories. 
While existing panoptic benchmarks~\cite{cao2024pasco, shi2024panossc} have significantly advanced scene understanding, they typically overlook the distinct instance boundaries of static environmental elements---such as buildings and poles---treating them instead as undifferentiated semantic masks. 
This oversight leads to a notable lack of instance-level coherence for the fixed environment, a fundamental limitation highlighted in recent urban mapping studies~\cite{li2025voxdet}.
In response, OccTrack360 introduces a unified 4D representation that extends identity consistency to all occupied voxels, encompassing both dynamic agents and static landmarks within a single temporally coherent framework.

\noindent\textbf{Fisheye-based Perception and Datasets.}
Fisheye cameras provide essential near-field surround-view coverage for autonomous systems, despite the challenges of severe radial distortion~\cite{qian2022survey}. 
Early research efforts primarily focused on 2D fisheye tasks, such as Bird’s-Eye View (BEV) segmentation~\cite{yogamani2024fisheyebevseg, liu2025articubevseg, wenke2025dur360bev, samani2023f2bev} and object detection~\cite{li2025exploring, kumar2021omnidet, wu2022surround_valet}, often employing specialized viewpoint augmentation or context-specific priors~\cite{cho2023surround_viewpoint_augmentation, pang2025dsbev}.
Recently, the scope has expanded toward 3D occupancy prediction, where distorted features are mapped into unified voxel or BEV spaces to resolve geometric ambiguities~\cite{wu2025omniocc, yang2025equivfisheye, li2025fishbev, sun2025kd360}. 
However, most existing benchmarks are limited to static semantic labeling or short-term sequences, lacking the temporally consistent instance IDs in 3D space required for long-term tracking. 
Consequently, as shown in Table~\ref{table:dataset_comparison}, OccTrack360 provides semantic and instantized occupancy annotations, facilitating evaluation of 4D panoptic occupancy tracking in fisheye-based systems. As shown in Fig.~\ref{fig:data_distribution}, the class distribution of OccTrack360 is generally aligned with SSCBench-KITTI360~\cite{li2024sscbench}, though notable discrepancies in truck and other-ground arise from differences in sensing range.

\section{OccTrack360: Established Benchmark}

We construct the OccTrack360 benchmark, based on KITTI‑360~\cite{Liao2021ARXIV}, to support 4D
panoptic occupancy tracking with voxel-level 3D object annotations. 
As illustrated in Fig.~\ref{fig:Fisheye_Benchmark_Pipeline}, our dataset construction pipeline consists of three stages: \textbf{(i) Voxel Generation}, which derives voxel annotations from visible fisheye regions. 
\textbf{(ii) Object Completion}, which reconstructs occluded and rear object parts for holistic 3D occupancy, and \textbf{(iii) Visibility Constraint}, which defines valid supervision regions through visibility and field-of-view masks. 
Compared with existing fisheye benchmarks~\cite{Liao2021ARXIV,yogamani2019woodscape,sekkat2022synwoodscape} that offer only sparse or instance-level labels, OccTrack360 provides \textbf{voxel-level} ground truth.

\begin{figure}[!t]
\centering
\includegraphics[width = 0.8\linewidth]{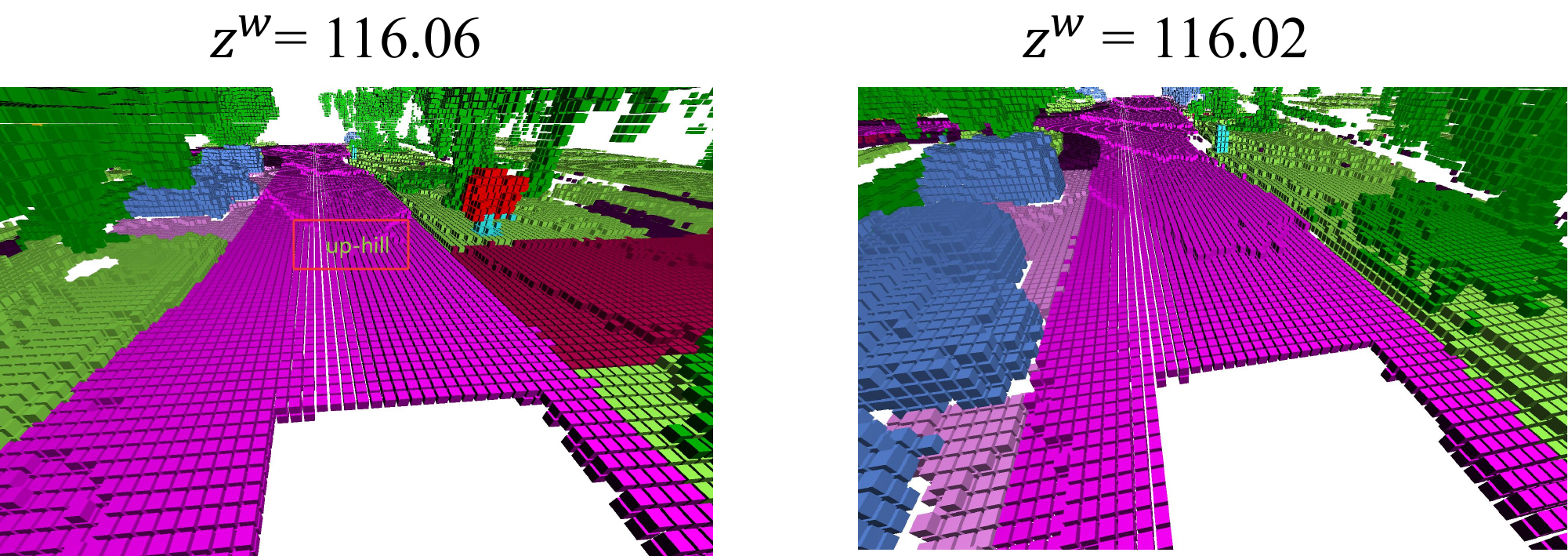}
\vskip-1ex
\caption{\textbf{Illustration of $z^+$ error.} 
The two frames shown above are both from sequence Seq00. 
The left and right subplots depict the $75\textsuperscript{th}$ and $100\textsuperscript{th}$ frames of SSCBench-KITTI360~\cite{li2024sscbench}, respectively, corresponding to the $167\textsuperscript{th}$ and $192\textsuperscript{th}$ frames of the KITTI360 dataset~\cite{Liao2021ARXIV}.
The value $z^w$ represents the height translation extracted from the extrinsic parameters provided by KITTI360~\cite{Liao2021ARXIV}. 
At the $75\textsuperscript{th}$ frame, the ground-truth voxel labels indicate the onset of an uphill segment. 
By the $100\textsuperscript{th}$ frame, the ego-vehicle has already ascended the slope, yet the corresponding $z^w$ value decreases.}
\label{fig:illustration_$z^+$_error}
\vskip-4ex
\end{figure}

\par
\textbf{Voxel Generation.} In the first stage, we follow KITTI-360~\cite{Liao2021ARXIV} and SSCBench-KITTI360~\cite{li2024sscbench} to project the bounding boxes with instance IDs into the ego coordinate system, denoted as $B^{ego}$. 
Each $B^{ego}$ is then matched to its corresponding semantic object and assigned the associated instance ID. To handle instance voxels outside the instance bounding boxes, we project them into the local coordinate system of each object $i$, denoted as $V^{i}$, select the subset with the minimum norm of $V^{i}$, and assign the corresponding instance ID.

\textbf{Object Completion.} After the above processing, the labels can be utilized for monocular TrackOcc. 
However, the SSCBench-KITTI360 labels contain only forward-facing voxels, 
making them unsuitable for a comprehensive understanding of the surrounding scene. To address this limitation, in the third stage, we reposition the ego vehicle from the back boundary to the center point. The completion of backward-facing voxels is then divided into two steps: static voxel alignment and dynamic voxel transformation.

In static voxel alignment, we align the current backward-facing voxels with the previous forward-facing voxels:
\begin{equation}
    V_{f_0}^{center} = V_{f_0}^{FF} + \sum_{i=1}^{N} \big( T^{ego}_{f_0 \leftarrow f_{0-i}} \cdot R \circ V_{f_{0-i}}^{FF} \big),
\end{equation}
where $f_0$ denotes the current frame and $f_{0-i}$ represents the historical frame at index $i$. $V_{f_0}^{FF}$ and $V_{f_{0-i}}^{FF}$ represent the forward-facing voxels at the corresponding frames, and $\circ$ denotes the Hadamard product. The term $T^{ego}_{f_0 \leftarrow f_{0-i}}$ denotes the ego-pose transformation matrix from frame $f_{0-i}$ to $f_0$. To maintain consistency with the Z-range ($[-2.0\,\text{m}, 4.4\,\text{m}]$) defined in SSCBench-KITTI360~\cite{li2024sscbench}, this transformation only considers rotation around the Z-axis and translation along the X and Y axes. The summation continues until the maximum depth of the projected voxels falls behind the rear boundary of the current frame.

To address this issue, we project the transformation onto the XY plane by isolating the rotation component around the $y$-axis together with the translations along the $x$- and $z$-axes. 
Singular Value Decomposition (SVD) is then performed for normalization, which suppresses errors but inevitably disregards genuine variations along the $z$-axis. 
Consequently, the matrix $R$ is designed to constrain the filling range of each frame, which is determined by the maximum depth of the previously occupied voxels at the current position.
In dynamic voxel transformation, for instance $I$, we first transformed the voxels of the previous $I$  to the current $I$ observed at the previous frame. 
Then, we transformed $I$ from the previous frame to the current frame.  
\begin{equation}
    V_{f_0}^{I, center} = V_{f_0}^{I,FF} + \sum_{i=1}^{N} \big( T^{ego}_{f_0 \leftarrow f_{0-i}} \cdot T_{f_0 \leftarrow f_{0-i}}^{I}  \cdot R \circ V_{f_{0-i}}^{FF}).
\end{equation}
The term $T^{ego}_{f_0 \leftarrow f_{0-i}}$ denotes the instance pose transformation matrix from frame $f_{0-i}$ to $f_0$.

\begin{figure}
\centering
\includegraphics[width=0.8\linewidth]{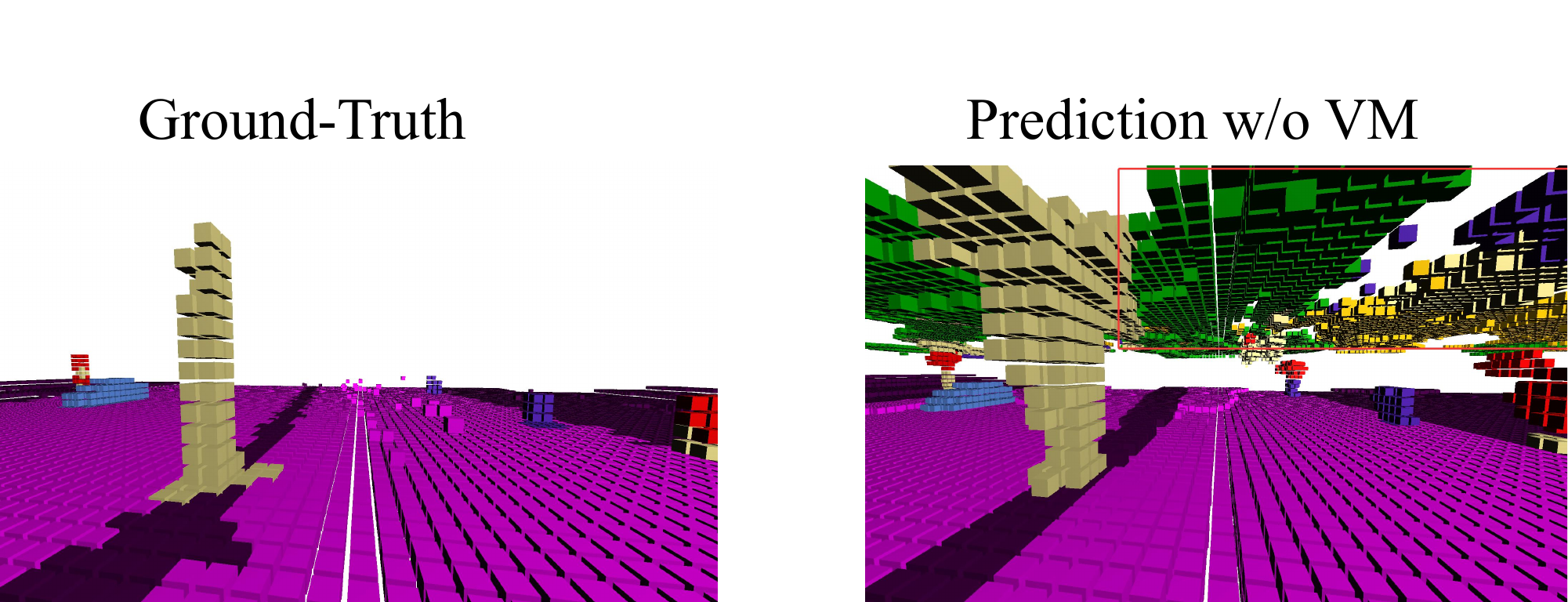}
\caption{\textbf{Limitation of the occlusion mask in Occ3D-Waymo}~\cite{tian2023occ3d}\textbf{.} 
VM denotes the occlusion mask. Predictions without the occlusion mask exhibit voxel ambiguity, as certain directions were filtered out during training and thus excluded from back-propagation.}
\label{fig:Limitation_of_the_occlusion_mask_in_Occ3D-Waymo}
\vskip-4ex
\end{figure}

\textbf{Visibility Constraint.} The occlusion mask is used to indicate which voxels are blocked, and it varies across frames as the surrounding voxels change.
The construction of the occlusion mask is primarily based on Occ3D~\cite{tian2023occ3d}. 
The key difference lies in the ray formation: Occ3D connects each occupied voxel center to the camera origin, whereas we connect each voxel located on the boundary of the voxel space.  
This discrepancy arises because certain directions—particularly upward ones—within the voxel space fail to connect to any occupied voxels, for which the occlusion mask in Occ3D-Waymo~\cite{tian2023occ3d} is incomplete and lacks coverage in those directions.  
This may result in filtering out more directions during training and inference, making the prediction visually and geometrically incomprehensible without the occlusion mask, as clearly shown in Fig.~\ref{fig:Limitation_of_the_occlusion_mask_in_Occ3D-Waymo}.

\begin{figure}[!t]
    \centering
    \includegraphics[width=0.95\linewidth]{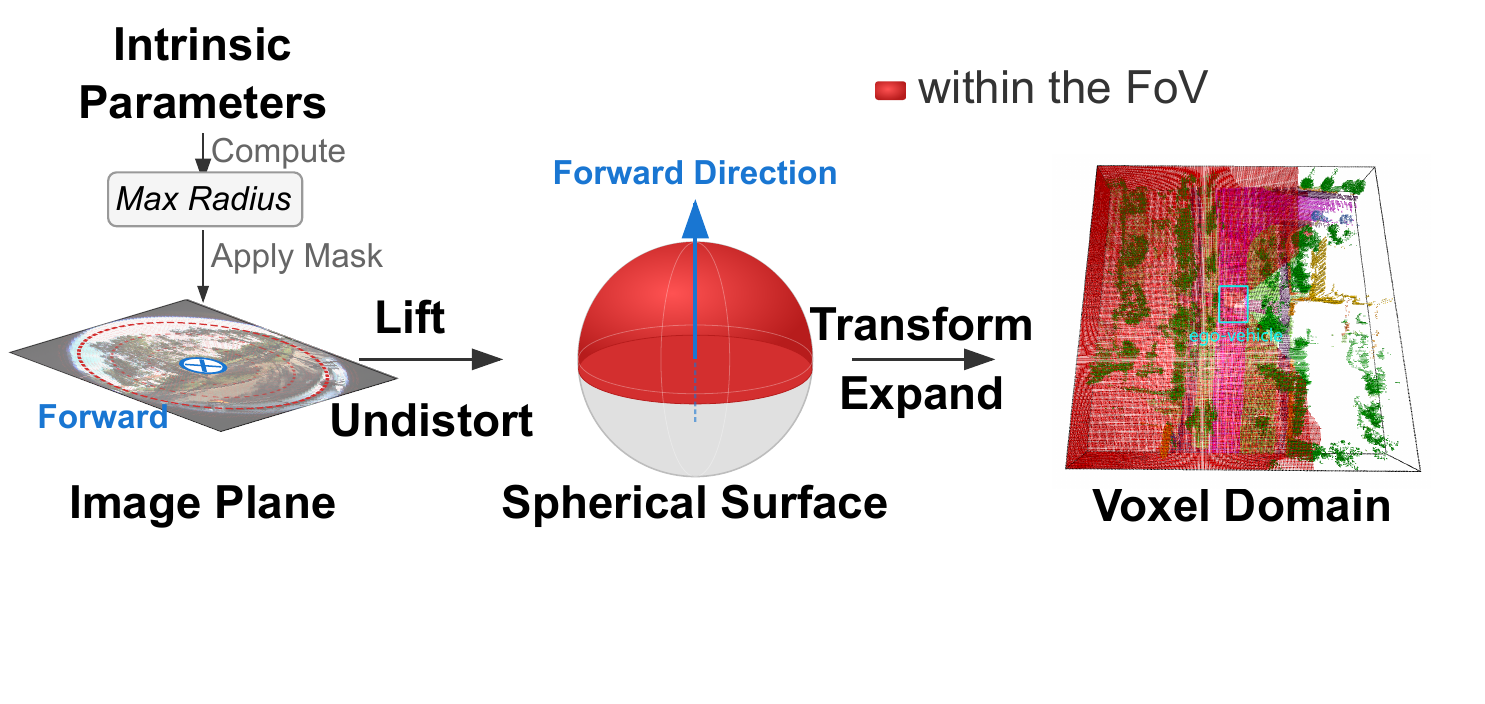}
    \vskip-7ex
    \caption{\textbf{Fisheye FoV mask pipeline.} 
    For illustration, we use the image plane captured by the ego vehicle’s left fisheye camera. 
    The maximum radius is first computed from the intrinsic parameters and projected onto the distorted image plane for rectification. The result is then lifted to a normalized spherical surface, followed by transformation according to the fisheye camera’s position and orientation. Finally, the spherical projection is extended to cover the entire voxel domain.}
    \label{fig:Fisheye_FoV_Mask_Pipeline}
    \vskip-2ex
\end{figure}

The fisheye FoV mask construction, as shown in Fig.~\ref{fig:Fisheye_FoV_Mask_Pipeline}, is based on the intrinsic parameters 
of the fisheye camera, calibrated using the Unified Projection Model (MEI)~\cite{mei2007single}. 
The mask is constructed in three steps. 

\begin{figure}[!t]
    \centering
    \includegraphics[width=0.25\textwidth]{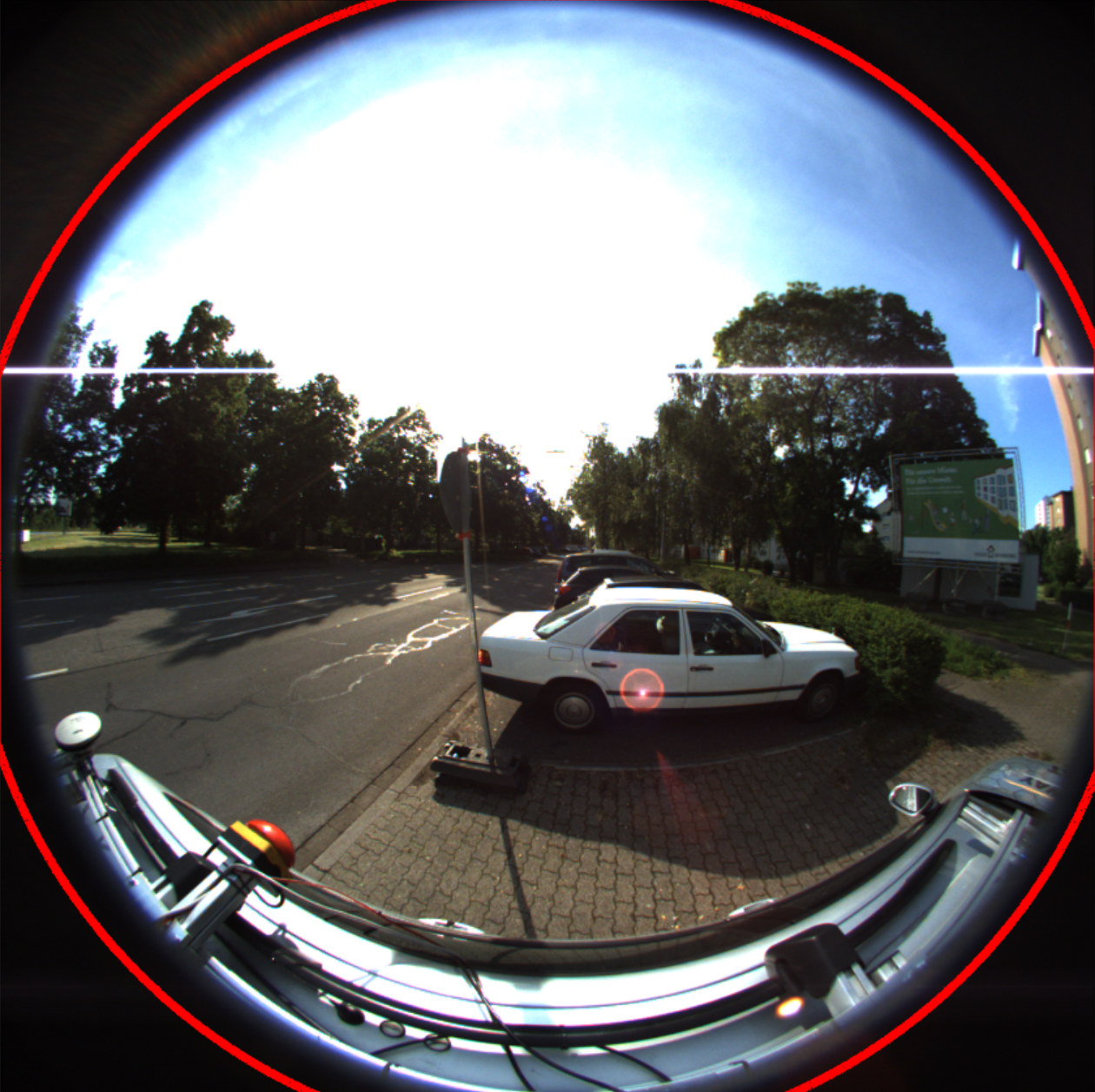}
    \vskip -1ex
    \caption{\textbf{$a'_{max}$ in distorted image-plane.}
    The red contour delineates the projection of the sampled radial parameter $a$ onto the distorted image plane, following the application of the distortion and intrinsic mapping functions.}
    \label{fig:a'_max in distorted image-plane}
    \vskip-4ex
\end{figure}

\begin{figure*}[!t]
\centering

\includegraphics[width = 0.95\textwidth]{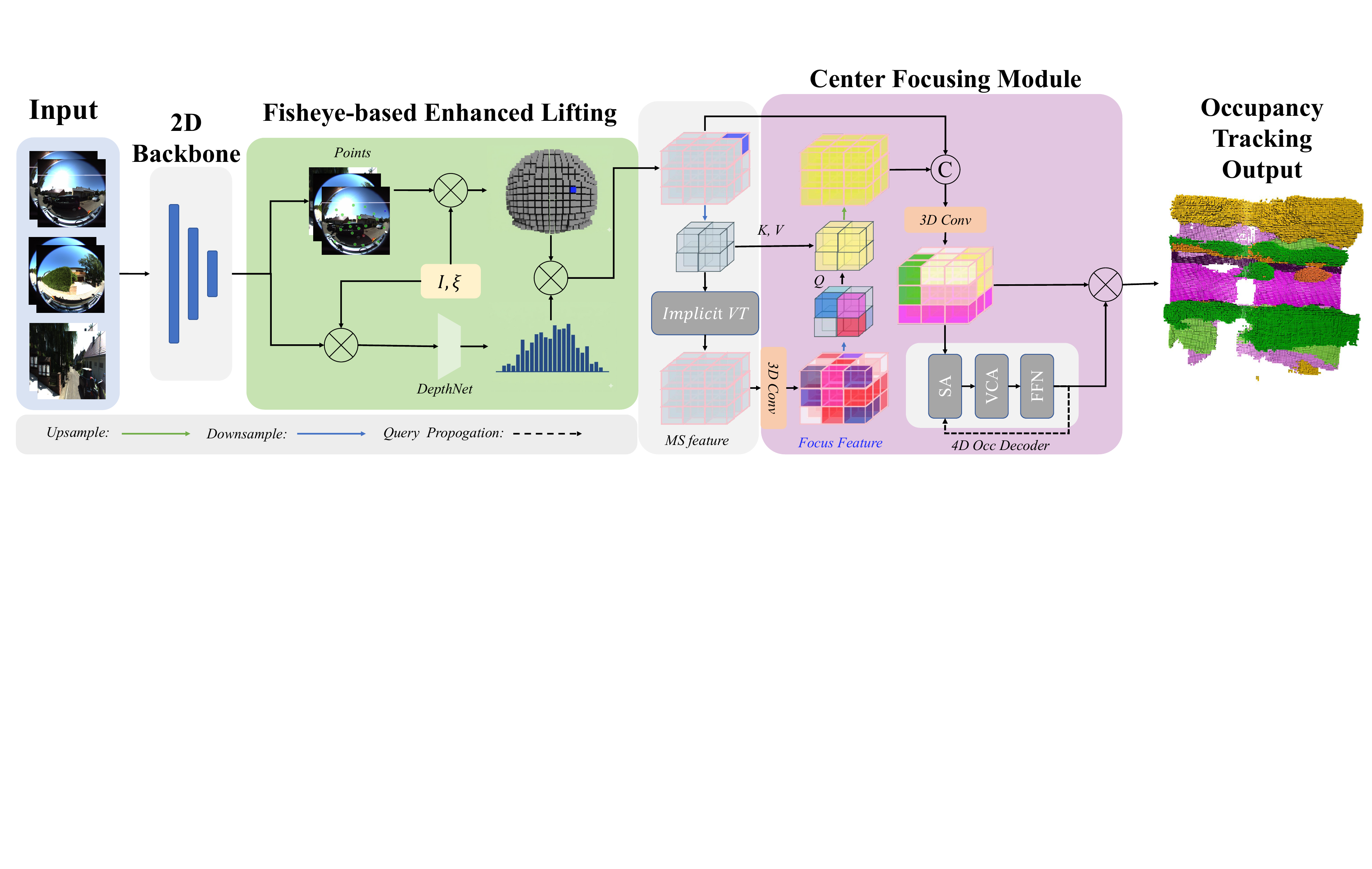}
\vskip-22ex
\vskip-11ex
\caption{\textbf{Overall pipeline of FoSOcc.} 
Our framework advances the standard occupancy pipeline through two novel contributions: 
(1) Fisheye-based Enhanced Lifting (FEL), which formulates the 2D-to-3D transformation within a spherical projection space by explicitly incorporating geometric priors—specifically the mirror parameter $\xi$, camera intrinsics, and distortion coefficients—to achieve high-fidelity spatial mapping; 
and (2) Center Focusing Module (CFM), which leverages multi-scale (MS) features from compact occupancy transformer representations~\cite{ma2024cotr} as inputs to facilitate precise object center localization via explicit supervision from 4D occupancy labels, thereby enhancing instance-level feature representation.}
\label{fig:overall_pipeline} 
\vskip-4ex
\end{figure*}

First, we determine the feasible projection limit of the undistorted normalized image-plane radius, denoted as $a_{\max}$. In spherical projection surface observed at shifted center, from camera per se, forward $z_{sh} = \cos\theta + \xi$, down $y_{sh} = \sin\theta  \sin\varphi$, right $x_{sh} = \sin\theta  \cos\varphi$, where the $\xi$ is the mirror parameter, the $\theta$ is the latitude of the unit sphere. 
The normalized radial ratio with the shifted center could be expressed as: 
\begin{equation}
a = \sqrt{\frac{{x_{sh}}^2 + {y_{sh}}^2}{z_{sh}^2}} = \frac{\sin\theta}{\cos\theta + \xi}.
\end{equation}
By squaring both sides and performing algebraic expansion, we obtain:
\begin{equation}
\cos\theta = \frac{-a^2\xi + \sqrt{\Delta}}{a^2 + 1},  \Delta = 1 - a^2(\xi^2 - 1).
\label{eq:cos_theta}
\end{equation}
Although the quadratic equation yields two possible solutions for $\cos\theta$, only the root associated with the physically valid projection is retained. 
In the unified camera model, this corresponds to the solution that ensures the ray intersects the spherical projection surface and can be mapped onto the image plane. 
While the negative root may mathematically exist, it represents rays that do not contribute to the effective field of view or lead to non-physical projections. In wide-FoV or fisheye configurations ($\xi > 1$), $\cos\theta$ can indeed take negative values; however, the chosen root is still the one consistent with the forward-facing imaging geometry defined by the model.
To obtain the real solution of $\cos\theta$,
\begin{equation}
    \Delta \geq 0 
\;\Longrightarrow\; 
a^2 \leq \frac{1}{\xi^2 - 1},a\geq0
\;\Longrightarrow\; 
a \leq \sqrt{\frac{1}{\xi^2 - 1}}.
\label{eq:real_solution}
\end{equation}
Therefore, the maximum feasible value is $a_{max} = \sqrt{\frac{1}{\xi^2 - 1}}$. 

Second, we sample $a \in [0,a_{max}]$, $\varphi \in [0, 2\pi]$ to compute the undistorted normalized plane values $x_u = acos\varphi$, $y_u = asin\varphi$. Subsequently, by applying the Brown-Conrady distortion~\cite{duane1971close} and the image‑plane mapping, the values are projected onto the distorted image plane with axis \textit{u} and \textit{v}.

As shown in Fig.~\ref{fig:a'_max in distorted image-plane}, the radial boundary \(a = a_{max}\) would ideally project to a circular contour on the image plane. 
In practice, however, the finite image resolution causes parts of this circle to extend beyond the image boundaries, 
particularly near the top \([0,h]\), bottom \([0,-h]\), left \([-w,0]\), and right \([w,0]\) edges, 
resulting in a truncated and non‑circular contour. 
Therefore, a manual truncation is required to restrict the contour within the valid image domain. 
To simulate the Field of View (FoV), the image-plane coordinates $(u,v)$ are restricted to the valid domain. Following the undistortion and inverse mapping operations, the rectified variable $a$ is derived and represented as $a'$. 
Substituting $a'$ for $a$ in Eq.~\eqref{eq:cos_theta} yields the corresponding value of $\theta$.

We then compute the coordinates ($x,y,z$) on the normalized spherical surface observed at the sphere center as follows:
\begin{equation}
    \begin{cases}
    x_{sp} = x_{sh} = sin\theta cos\varphi,\\
    y_{sp} = y_{sh} = sin\theta sin\varphi,\\
    z_{sp} = z_{sh} - \xi
 = cos\theta.\end{cases}
 \label{eq:position in spherical center}
\end{equation}
where $x_{sp}$ is the $x$-component relative to the sphere center, while $x_{sh}$ is the $x$-component relative to the shifted center. 
Finally, the normalized spherical surface is extended across the entire voxel domain, thereby establishing the feasible representation. Building on the previously obtained normalized spherical surface, the 3D points are transformed using the camera installation pose provided in the KITTI360 dataset~\cite{Liao2021ARXIV}.

These points are then interpreted as direction vectors and extended to populate the entire voxel space.

\section{FoSOcc: Proposed Method}
\begin{figure*}[!t]
    \centering
    \includegraphics[width=0.9\linewidth]{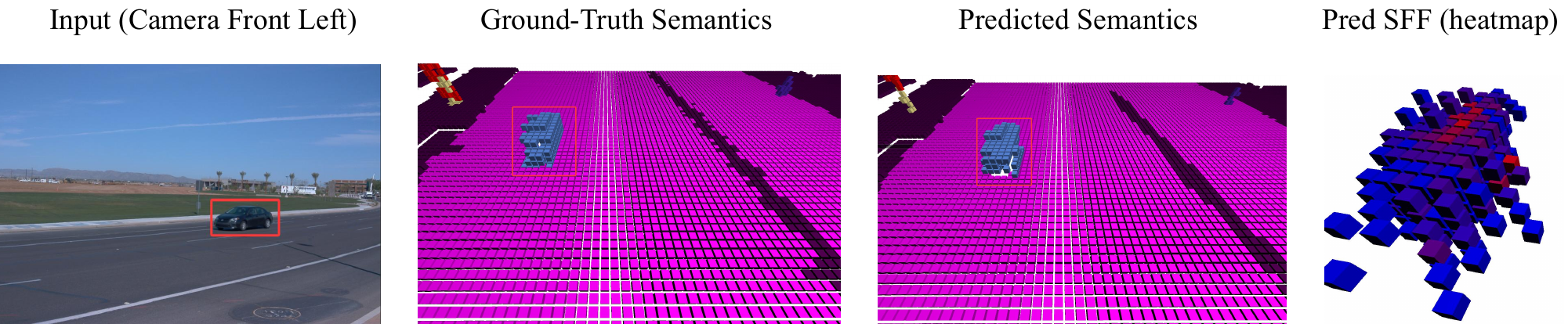}
    \vskip -1ex
    \caption{\textbf{Visualization of the Supervised Focus Feature (SFF).} 
    The heatmap provides a magnified view of the predicted semantic region. 
    In the heatmap, red voxels represent higher digital values compared to the blue regions.}
    \label{fig:Visualization_Supervised_Focus_Feature}
    \vskip-3ex
\end{figure*}

\subsection{Overall Architecture}
To mitigate the geometric distortions and non-uniform resolution inherent in fisheye optics, our proposed occupancy prediction pipeline (Fig.~\ref{fig:overall_pipeline}) integrates two synergistic components: (i) Center Focusing Module (CFM), which reinforces instance-level feature aggregation by supervising object centroids, and (ii) a Fisheye-based Enhanced Lifting (FEL), designed to achieve robust 2D-to-3D feature transformation by explicitly modeling spherical geometry.

\subsection{Center Focusing Module}
To capture fine-grained instance geometry and improve localization robustness under distorted projections, we propose the \textit{Center Focusing Module} (CFM). This module generates dense geometric cues to supervise the representation of object interiors and centroids.

\subsubsection{Revisiting Voxel Offsets}
The previous work~\cite{li2025voxdet} introduced a voxel offset module to encode boundary details. 
The ground-truth offset for a voxel $(i, j, k)$ is defined by the distance to the instance boundary along six cardinal directions ($x^\pm, y^\pm, z^\pm$). Taking the positive x-direction ($x^+$) as an example:
\begin{equation}
\delta^{x^+}_{i,j,k} =
\begin{cases}
\delta^{x^+}_{i-1,j,k} + 1, & \text{if } sem[i-1,j,k] = sem[i,j,k], \\
1, & \text{otherwise or } i-1 = 0.
\end{cases}
\label{eq:delta}
\end{equation}
where $sem[i, j, k]$ represents the ground-truth semantic occupancy label at the voxel position $(i, j, k)$.
Subsequently, for each semantic class $c$, the offsets are normalized by dividing them by the shape of the voxel space. This normalization facilitates stable training and accelerates model convergence.
Yet, the existing voxel offset suffers from two drawbacks. 

\noindent i) Sensitivity to depth inaccuracy: Standard offsets peak at object boundaries. In fisheye optics, peripheral regions suffer from severe tangential distortion and non-uniform resolution, leading to inevitable errors in 2D-to-3D depth lifting. Even a single-voxel misalignment in depth can cause the loss function to penalize the model with the maximum possible error, \textit{i.e.}, $\mathcal{L}_{1}\!\left(\mathbf{0}^{\text{gt}}, \mathbf{Max}^{\text{pred}}\right)$, leading to unstable gradients.

\noindent ii) Geometric ambiguity in global normalization: Conventional methods normalize offsets by the entire voxel grid dimensions. 
Since semantic classes (\textit{e.g.}, buildings \textit{vs.} pedestrians) vary drastically in scale, a fixed normalization factor renders the offset values physically ambiguous, making it difficult for the network to learn a universal geometric representation across different categories.

\subsubsection{Formulation of CFM}
To address these issues, the Center Focusing Module (CFM) shifts the supervisory focus from volatile boundaries to stable instance centers. We define a product-based focus feature $\delta^p$ by aggregating offsets from all six directions:
\begin{equation}
\delta^p_{i,j,k} = \prod_{d \in {x^\pm, y^\pm, z^\pm}} \delta^{d,\text{norm}}_{i,j,k}.
\end{equation}
By construction, $\delta^p{i,j,k}$ naturally reaches its \textbf{maximum at the geometric center} of an instance and decays toward the boundaries. This ``center-peaked'' distribution acts as a Gaussian-like soft constraint, which is far more tolerant to the spatial jitter caused by fisheye distortion than hard boundary offsets.

Leveraging instance-level annotations, we normalize the $\delta^p$ and individual offsets using the maximum value within each specific instance $I$:
\begin{equation}
\delta^{p,\text{norm},I}_{i,j,k} = \frac{\delta^{p, I}_{i,j,k}}{\max(\delta^{p, I}_{i,j,k} + \epsilon)}.
\end{equation}
This ensures that the geometric cues remain scale-invariant, allowing the model to learn consistent internal structures regardless of whether the object is a large background structure or a small dynamic agent. 
As illustrated in Fig.~\ref{fig:Visualization_Supervised_Focus_Feature}, the SFF provides a spatially smooth supervisory signal that anchors the model's attention on stable instance centers, thereby maintaining localization robustness even in distorted peripheral regions where depth-based lifting is imprecise.

\subsection{Fisheye-based Enhanced Lifting}

Conventional 2D-to-3D lifting frameworks~\cite{philion2020lift} typically assume a pinhole camera model, relying on linear intrinsic parameters to project pixels into 3D space. 
However, such models fail to account for the severe non-linear radial distortions and wide-FoV characteristic of fisheye optics. 
To bridge this geometric gap, we propose \textit{Fisheye-based Enhanced Lifting} (FEL), which reformulates the lifting process using the Unified Projection Model (MEI)~\cite{mei2007single}. 
By incorporating a mirror parameter $\xi$, FEL explicitly models the projection onto a displaced unit sphere, providing a more expressive geometric foundation for wide-FoV occupancy prediction.

The transformation pipeline begins with the normalized image coordinates $\mathbf{p} {=} [x, y]^\top$. 
Given the corresponding depth bins $D$, we first represent the point in polar form $(a, \varphi)$, where $a {=} \sqrt{x^2+y^2}$ and $\varphi {=} \operatorname{atan2}(y, x)$.
To ensure the geometric validity of the projection under non-pinhole configurations (\textit{i.e.}, $\xi \ne 0$), 
we apply Eq.~\eqref{eq:real_solution} for rectification constraint on the radial distance $a$. 
Subsequently, the incident angle $\theta$ is derived from the rectified $a$ via the cosine relation defined in Eq.~\eqref{eq:cos_theta}. 
By combining these angular components, we recover the 3D coordinate on the unit sphere according to Eq.~\eqref{eq:position in spherical center}.
Finally, the resulting unit vectors are scaled by the discretized depth bins $D$ to lift the features into the 3D spatial manifold, ensuring that each voxel accurately aligns with its physical light ray in the fisheye coordinate system.

\section{Experiments}

\begin{table*}[!t]
\centering
\caption{\textbf{Results on Occ3D-Waymo.} \textit{G.O.} denotes general objects. Methods are evaluated on categories with and without instance IDs. Best results are \textbf{bolded}, and ours is highlighted in green.}
\vskip-1ex
\resizebox{\textwidth}{!}{
\begin{tabular}{l|c|ccccc|cccc|c} %
\toprule
 &  & \multicolumn{5}{c|}{OccSQ} & \multicolumn{4}{c|}{OccAQ} &  \\ \cline{3-11} 
\multirow{-2}{*}{Method} & \multirow{-2}{*}{E2E} & \multicolumn{1}{c|}{Overall} & Sign~\textcolor{sign_c}{\rule{0.2cm}{0.2cm}} & Building~\textcolor{building_c}{\rule{0.2cm}{0.2cm}} & Vegetation~\textcolor{vegetation_c}{\rule{0.2cm}{0.2cm}} & \textit{G.O.}~\textcolor{go_c}{\rule{0.2cm}{0.2cm}} & \multicolumn{1}{c|}{Overall} & Vehicle~\textcolor{vehicle_c}{\rule{0.2cm}{0.2cm}} & Pedestrian~\textcolor{pedestrian_c}{\rule{0.2cm}{0.2cm}} & Cyclist~\textcolor{cyclist_c}{\rule{0.2cm}{0.2cm}} & \multirow{-2}{*}{OccSTQ} \\ 
\midrule
MinVIS~\cite{huang2022minvis} & \ding{51} & \multicolumn{1}{c|}{29.1} & 10.6 & 39.1 & 34.0 & \textbf{11.7} & \multicolumn{1}{c|}{3.3} & 4.0 & 1.5 & 3.1 & 9.8 \\
CTVIS~\cite{ying2023ctvis} & \ding{51} & \multicolumn{1}{c|}{26.5} & 13.2 & 36.9 & 30.9 & 8.1 & \multicolumn{1}{c|}{4.3} & 5.3 & 2.1 & 1.6 & 10.7 \\
AB3DMOT~\cite{weng20203d} & \ding{55} & \multicolumn{1}{c|}{29.1} & 10.7 & 39.2 & 34.0 & \textbf{11.7} & \multicolumn{1}{c|}{5.3} & 6.8 & 1.7 & 2.9 & 12.4 \\
4D-LCA~\cite{aygun20214d} & \ding{55} & \multicolumn{1}{c|}{29.1} & 10.7 & 39.2 & 34.0 & \textbf{11.7} & \multicolumn{1}{c|}{9.0} & 11.4 & \textbf{3.4} & 4.1 & 16.2 \\
TrackOcc (baseline)~\cite{chen2025trackocc} & \ding{51} & \multicolumn{1}{c|}{29.4} & 13.6 & 43.2 & 40.0 & 9.2 & \multicolumn{1}{c|}{13.5} & 18.1 & 3.1 & 4.6 & 20.0 \\
\rowcolor{ourgreen}
FoSOcc (Ours) & \ding{51} & \multicolumn{1}{c|}{\cellcolor{ourgreen}\textbf{30.8}} & \textbf{15.3} & \textbf{44.9} & \textbf{40.8} & 11.1 & \multicolumn{1}{c|}{\cellcolor{ourgreen}\textbf{14.2}} & \textbf{18.9} & 3.2 & \textbf{5.8} & \textbf{20.9} \\
\bottomrule
\end{tabular}
}
\label{tab:occ3d_waymo}
\vskip-2ex
\end{table*}

\begin{table*}
\centering
\caption{\textbf{Impact of input modalities on OccTrack360.} The ``all'' setting utilizes the FoV, while the ``Rectified Fisheyes'' and ``Fisheyes'' settings are constrained to the fisheye-specific FoV. Abbreviations: O.V. (Other Vehicle), O.S. (Other Structure).}
\resizebox{0.9\textwidth}{!}{
\begin{tabular}{lc|ccccc|cccc|c}
\toprule
\multirow{2}{*}{Method} & \multirow{2}{*}{Inputs \& FoV} & \multicolumn{5}{c|}{OccSQ} & \multicolumn{4}{c|}{OccAQ} & \multirow{2}{*}{OccSTQ} \\ 
\cline{3-11}
& & \multicolumn{1}{c|}{Overall} & parking~\textcolor{parking_c}{\rule{0.2cm}{0.2cm}} & Building~\textcolor{building_c}{\rule{0.2cm}{0.2cm}} & Fence~\textcolor{fence_c}{\rule{0.2cm}{0.2cm}} & O.S.~\textcolor{os_c}{\rule{0.2cm}{0.2cm}} & Overall & Car~\textcolor{vehicle_c}{\rule{0.2cm}{0.2cm}} & Truck~\textcolor{truck_c}{\rule{0.2cm}{0.2cm}} & O.V.~\textcolor{ov_c}{\rule{0.2cm}{0.2cm}} & \\
\midrule
TrackOcc & \multirow{2}{*}{\textit{all}} & \multicolumn{1}{c|}{12.90} & 0 & 29.87 & 0.85 & 7.02 & 17.07 & 20.69 & 8.38 & 3.40 & 14.84 \\
\cellcolor{ourgreen}FoSOcc (Ours) & & \multicolumn{1}{c|}{\cellcolor{ourgreen}{14.20}} & \cellcolor{ourgreen}7.56 & \cellcolor{ourgreen}25.09 & \cellcolor{ourgreen}2.90 & \cellcolor{ourgreen}12.70 & \cellcolor{ourgreen}17.61 & \cellcolor{ourgreen}21.53 & \cellcolor{ourgreen}9.87 & \cellcolor{ourgreen}5.03 & \cellcolor{ourgreen}15.82 \\ 
\hline
TrackOcc & \textit{Rectified Fisheyes} & \multicolumn{1}{c|}{10.15} & 5.26 & 17.44 & 2.91 & 8.52 & 12.97 & 15.62 & 5.80 & 4.11 & 11.48 \\
\hline
TrackOcc & \multirow{2}{*}{\textit{Fisheyes}} & \multicolumn{1}{c|}{1.59} & 1.02 & 4.37 & 0.34 & 0.03 & 1.76 & 2.16 & 0.74 & 0.14 & 1.67 \\
\cellcolor{ourgreen}FoSOcc (Ours) & & \multicolumn{1}{c|}{\cellcolor{ourgreen}14.37} & \cellcolor{ourgreen}6.33 & \cellcolor{ourgreen}33.75 & \cellcolor{ourgreen}3.73 & \cellcolor{ourgreen}12.95 & \cellcolor{ourgreen}14.40 & \cellcolor{ourgreen}17.63 & \cellcolor{ourgreen}6.64 & \cellcolor{ourgreen}1.69 & \cellcolor{ourgreen}14.38 \\ 
\bottomrule
\end{tabular}
}
\label{table:different inputs in OccTrack360}
\end{table*}

\subsection{Datasets and Evaluation Metrics}
To comprehensively evaluate our proposed approach, we perform experiments and ablation studies on two benchmarks: \textbf{Occ3D-Waymo}~\cite{tian2023occ3d} and the established \textbf{OccTrack360}. 
In \textbf{Occ3D-Waymo}~\cite{tian2023occ3d}, the spatial range along the forward-backward and left-right directions extends from \([{-}40m,\,40m]\), while the vertical (up-down) range spans \([{-}1m,\,5.4m]\). 
Each voxel corresponds to a resolution of \(0.4m{\times}0.4m{\times}0.4m\), 
and the input data are provided at a frequency of \(2Hz\).
In the proposed \textbf{OccTrack360}, the spatial coverage is defined over \([{-}25.6m,25.6m]\) in both the forward-backward and left-right directions, and \([{-}2.0m,4.4m]\) along the \(z\) (vertical) axis. Each voxel has a finer resolution of \(0.2m{\times}0.2m {\times}0.2m\), and the input data are also provided at \(2Hz\). 
To accommodate limited GPU memory, however, we restrict the effective range in our experiments to \([{-}12.8m,12.8m]\) in the horizontal plane. 
Following~\cite{chen2025trackocc}, we adopt Occupancy Segmentation and Tracking Quality (OccSTQ), Occupancy Segmentation Quality (OccSQ), and Occupancy Association Quality (OccAQ) as the primary metrics for evaluating 4D occupancy tracking.

\begin{table*}[!t]
\centering
\caption{\textbf{Ablations of each adjustment in the CFM on Occ3D-Waymo.} Here, \textit{A} and \textit{B} denote \textit{instance-level normalization} and \textit{supervised focus feature}, respectively. \textit{G.O. }denotes general objects. \textit{C.C.} denotes construction cones.
All configs except \emph{full-train} are trained for half of the original epochs ($12$ out of $24$).
Best results in $12$ epochs are \textbf{bolded}.}
\label{table:cfm_ablation_in_waymo}
\vskip -1ex
\resizebox{0.9\textwidth}{!}{
\begin{tabular}{cc|cccccc|cccc}
\toprule
\multirow{2}{*}{A} & \multirow{2}{*}{B} & \multicolumn{6}{c|}{OccSQ} & \multicolumn{4}{c}{OccAQ} \\ 
\cline{3-12}
 & & \multicolumn{1}{c|}{Overall} & Sign~\textcolor{sign_c}{\rule{0.2cm}{0.2cm}} & Building~\textcolor{building_c}{\rule{0.2cm}{0.2cm}} & Vegetation~\textcolor{vegetation_c}{\rule{0.2cm}{0.2cm}} & \textit{G.O.}~\textcolor{go_c}{\rule{0.2cm}{0.2cm}} & \textit{C.C.}~\textcolor{cc_c}{\rule{0.2cm}{0.2cm}} & \multicolumn{1}{c|}{Overall} & Vehicle~\textcolor{vehicle_c}{\rule{0.2cm}{0.2cm}} & Pedestrian~\textcolor{pedestrian_c}{\rule{0.2cm}{0.2cm}} & Cyclist~\textcolor{cyclist_c}{\rule{0.2cm}{0.2cm}} \\ 
\midrule
\ding{55} & \ding{55} & \multicolumn{1}{c|}{29.99} & 14.96 & 44.00 & 40.46 & 11.78 & 15.51 & \multicolumn{1}{c|}{10.62} & 14.02 & 2.68 & 5.43 \\
\ding{51} & \ding{55} & \multicolumn{1}{c|}{30.16} & 14.27 & 44.27 & 40.03 & 12.57 & 17.88 & \multicolumn{1}{c|}{\textbf{11.21}} & \textbf{14.89} & 2.71 & 4.91 \\
\ding{51} & \ding{51} & \multicolumn{1}{c|}{\textbf{30.67}} & \textbf{16.22} & \textbf{45.12} & \textbf{40.71} & \textbf{12.92} & \textbf{20.15} & \multicolumn{1}{c|}{10.65} & 14.06 & \textbf{2.72} & \textbf{5.73} \\ \hline
\rowcolor{ourgreen}
\multicolumn{2}{c|}{\emph{full-train}} & \multicolumn{1}{c|}{30.80} & 15.28 & 44.87 & 40.79 & 11.13 & 19.62 & \multicolumn{1}{c|}{14.18} & 18.90 & 3.19 & 5.79 \\ 
\bottomrule
\end{tabular}
}
\vskip-4ex
\end{table*}

\subsection{Implementation Details}
We apply a single-layer cross-attention to connect the outputs of {CFM} to 3D features. 
Following~\cite{chen2025trackocc}, each frame is initialized with $200$ emerging queries, with the query dimension consistently set to $256$ both before and after the update. 
The query lifecycle is governed by $\tau_1{=}0.3$, $\tau_2{=}0.25$, and $T_f {=}3$. 
For the visual front-end, we leverage ResNet-50 as the backbone to extract multi-view image features.
Our primary experiments are conducted after $24$ epochs of training with a learning rate of $2e{-}4$, using the AdamW optimizer with a total batch size of $4$, on four NVIDIA GeForce RTX 3090 GPUs. 

For evaluation on Occ3D-Waymo~\cite{tian2023occ3d}, we employ the 2D-to-3D lifting paradigm established in~\cite{philion2020lift}.
Furthermore, for evaluation on the established OccTrack360 benchmark, we incorporate our proposed FEL as the 2D-to-3D lifting mechanism. 
Only classes $1{\sim}6$ are considered for tracking evaluation. 
Inference experiments are conducted on a single NVIDIA GeForce RTX 3090 GPU, with evaluations specifically performed on the Seq06 dataset. 
All image resolutions are resized to $704{\times}704$.

\subsection{Comparison with State-of-the-Art on Occ3D-Waymo}
The 4D panoptic occupancy tracking results for the Occ3D-Waymo dataset~\cite{tian2023occ3d} are summarized in Table~\ref{tab:occ3d_waymo}. Our proposed CFM significantly outperforms the baseline, achieving relative SQ improvements of $11.1\%$ for traffic signs and $20.7\%$ for general objects. 
Furthermore, a substantial AQ gain of $26.1\%$ is observed in the cyclist category. 
By leveraging the CFM, our framework demonstrates enhanced adaptability in perceiving small-scale categories.

\subsection{Comparison with State-of-the-Art on OccTrack360}
Table~\ref{table:different inputs in OccTrack360} summarizes the performance comparison on the OccTrack360 dataset. Our proposed method consistently outperforms TrackOcc~\cite{chen2025trackocc}, achieving superior OccSTQ, OccAQ, and OccSQ scores across both \textit{Fisheyes-only} and \textit{All-input} configurations. Notably, while TrackOcc~\cite{chen2025trackocc} relies on rectifying fisheye images into perspective views—a process that inherently limits the field-of-view (FoV) and degrades coverage—our FoSOcc framework processes raw fisheye inputs directly, thereby preserving a more comprehensive sensing range.

\subsection{Ablation Studies}
Table~\ref{table:cfm_ablation_in_waymo} illustrates the efficiency of the Center Focusing Module (CFM). 
Specifically, the \textit{instance-level normalization} of ground-truth offsets improves the ability to capture geometric details of relatively large objects, such as vehicles. 
The \textit{supervised focus feature} consistently localizes its peak response at the geometric center of each instance, leading to improved localization accuracy for small-scale objects. Consequently, a significant SQ boost is observed in small-category instances, such as construction cones and signs.
The full CFM effectively combines both strategies, leveraging their complementary strengths. This integration enables more accurate and robust object-centric occupancy estimation across a wide range of scales and motion dynamics. As a result, CFM achieves consistent performance gains over baseline variants, demonstrating its effectiveness in enhancing 3D scene understanding, especially in complex, real-world autonomous driving scenarios.

\section{Conclusion}
In this paper, we look into 4D panoptic occupancy tracking from surround-view fisheye cameras.
To facilitate a comprehensive and wide Field-of-View (FoV) spatiotemporal surrounding understanding, we establish the OccTrack360 benchmark for omnidirectional 4D panoptic occupancy tracking, which provides long-term and diverse sequences with an all-directional occlusion mask and an MEI-based fisheye FoV mask. 
To develop a fisheye-oriented tracking baseline, we propose Focus on Sphere Occ (FoSOcc), which tackles distorted spherical projection and inaccurate voxel-space localization via novel center focusing and spherical lift modules.
Extensive experiments show FoSOcc achieves leading 4D panoptic occupancy tracking performance on Occ3D-Waymo and the established OccTrack360 benchmarks.

\bibliographystyle{IEEEtran}
\bibliography{bib}

\end{document}